\documentclass[letterpaper,10pt,conference,final]{ieeeconf}

\IEEEoverridecommandlockouts
\usepackage{graphics} 
\usepackage{mathptmx} 
\usepackage{amsmath} 
\usepackage{amssymb} 
\usepackage{cite}
\usepackage{enumerate}
\usepackage{xcolor}
\usepackage{graphicx}
\usepackage{amsfonts}
\usepackage{mathtools}
\usepackage[mathcal]{euscript}
\usepackage{caption}
\usepackage[hidelinks]{hyperref}
\usepackage[T1]{fontenc} 
\usepackage[hyphenbreaks]{breakurl} 
\usepackage{subcaption}
\usepackage[ruled,vlined]{algorithm2e}

\DeclareCaptionLabelSeparator{periodspace}{.\quad}
\usepackage{amsthm}

\theoremstyle{definition} 

\title{\LARGE \bf
	Robust 3D Distributed Formation Control  \\ with Application to Quadrotors
}

\author{Kaveh Fathian, Sleiman Safaoui, Tyler H. Summers, Nicholas R. Gans
	\thanks{*This work was supported by the U.S. Air Force Research Laboratory under grant FA8651-17-1-0001 and the Army Research Office under grant W911NF-17-1-0058.}
	\thanks{K. Fathian, S. Safaoui, and N. R. Gans are with the Department of Electrical Engineering, T. H. Summers is with the  Department of Mechanical Engineering, University of Texas at Dallas, Richardson, TX, 75080 USA.~  E-mail: {\tt\small \{kaveh.fathian, sxs169833, tyler.summers, ngans\}@utdallas.edu}        }%
}

\def\bn{\mathbb N}

\def\br{\mathbb R}

\newcommand{\bbar}[1]{\bar{\bar{#1}}}

\begin{document}

\maketitle
\thispagestyle{empty}
\pagestyle{empty}

\begin{abstract}
We present a distributed control strategy for a team of quadrotors to autonomously achieve a desired 3D formation. Our approach is based on local relative position measurements and does not require global position information or inter-vehicle communication. 
We assume that quadrotors have a common sense of direction, which is chosen as the direction of gravitational force measured by their onboard IMU sensors.
However, this assumption is not crucial, and our approach is robust to  inaccuracies and effects of acceleration on gravitational measurements.
In particular, converge to the desired formation is unaffected if each quadrotor has a velocity vector that projects positively onto the desired velocity vector provided by the formation control strategy.  
We demonstrate the validity of proposed approach in an experimental setup and show that a team of quadrotors achieve a desired 3D formation.
\end{abstract}

\section*{Supplementary Material}

Video of the experiments is available at 
{\color{blue} \href{https://youtu.be/KDA4XR2yoeI}{https://youtu.be/KDA4XR2yoeI}}.

\section{Introduction}

Thanks to technological advances in recent years, it is now possible to deploy a team of unmanned aerial vehicles to collaboratively map and monitor an environment  \cite{Cieslewski2018}, inspect infrastructures \cite{Ozaslan2017}, deliver goods \cite{Dorling2017}, or manipulate objects \cite{Baehnemann2017}.
In these applications, the ability to bring the vehicles to a desired geometric shape (i.e., formation) is a fundamental building block upon which more sophisticated maneuvering and navigation policies can be constructed.

There exists a large body of work on formation control of autonomous vehicles \cite{Oh2015}. However, many methods rely on a centralized motion planning scheme or a global positioning/communication paradigm \cite{Michael2008, Aranda2015, Hyun2016}.
Fully distributed formation control strategies \cite{Trinh2018, Fathian2016, Fathian2016a, Han2018}, on the other hand, do not have these requirements and compared to the centralized methods have better scalability, naturally parallelized computation, and resiliency to global positioning signal jamming or loss.
%

\begin{figure} [t!]
	\begin{center}
		\includegraphics[trim =20mm 10mm 20mm 1mm, clip, width=0.40\textwidth]{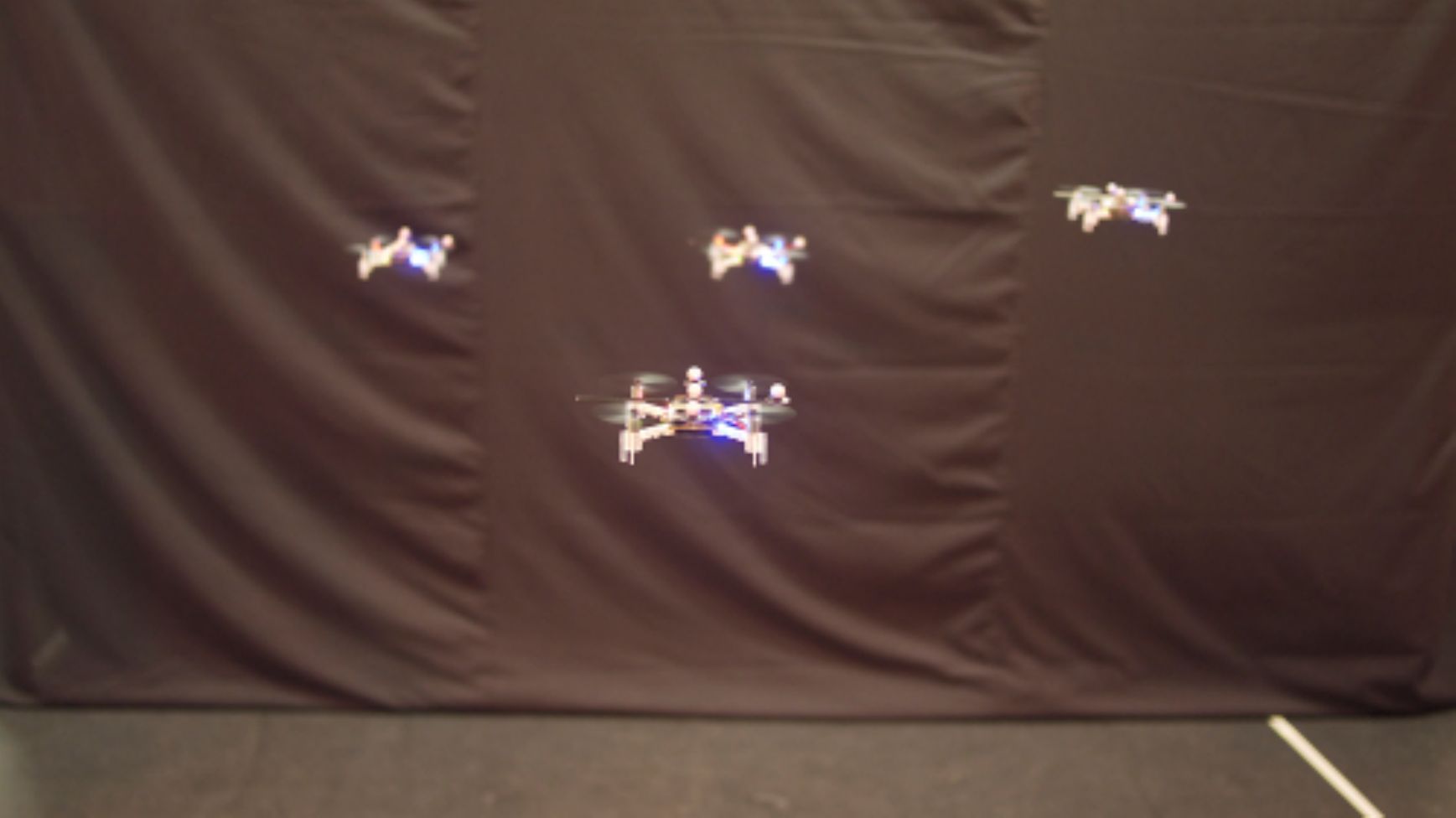}	
		\caption{Crazyflie quadrotors achieving a 3D formation.}
		\label{fig:Quads}
	\end{center}
\end{figure}

In this abstract, we present a distributed control strategy for a team of quadrotors to achieve a desired 3D formation. 
We assume that quadrotors have a common sense of direction, which is chosen as the direction of gravitational force measured by their onboard IMU sensors and is used to align the $z$-axes of their local coordinate frames.
The local relative positions of neighboring vehicles are multiplied by a set of constant gain matrices that are provided to each quadrotor before the mission, and summed to derive a desired direction of motion.

The main contribution of this abstract include extension of our previous work on planar formations \cite{Fathian2018, Fathian2018b} to 3D formations and a novel approach to design the required control gain matrices via semidefinite programming. 
Our approach can be applied on existing quadrotor platforms and is robust to noise, disturbances, and forces that can affect the estimated direction of gravity\footnote{In particular, our approach is robust to accelerometer measurement inaccuracies discussed in \cite{Leishman2014}.}.  
Furthermore, no global position information is needed, and a distributed collision avoidance scheme naturally arises from our  strategy.

\section{Methodology} \label{sec:Methodology}

In this section, we formulate the control strategy followed by a brief discussion on its design and properties. Due to space constraints and  nature of this manuscript, mathematical proofs are omitted and will be presented in a future work.

\subsection{Distributed Formation Control} 

To bring a team of $n \in \bn$ quadrotors to a desired 3D formation, we define the desired direction of motion $u_i \in \br^{3}$ for each quadrotor as
%
\begin{gather} \label{eq:HolonomCtrl}
u_i := \sum_{j \in \mathcal{N}_i}{A_{ij} \, q_j^{\,i} },
\end{gather}
%
where $i \in \{1,\, 2,\, \dots,\, n \}$ denotes the index of each quadrotor, $\mathcal{N}_i \subset \bn$ is the index set of neighboring vehicles for the $i$-th quadrotor, $q_j^{\,i} \in \br^{3}$ denotes the position of quadrotor $j$ in quadrotor $i$'s local coordinate frame, and $A_{ij}  \in \br^{3\times 3}$ are constant control gain matrices that have the form 
%
\begin{equation} \label{eq:Aij}
A_{ij} = \begin{bsmallmatrix}
a_{ij} & -b_{ij} & 0 \\
b_{ij} & a_{ij} & 0 \\
0 & 0 & c_{ij} \\
\end{bsmallmatrix},
\end{equation}
%
and are provided to quadrotors before the mission. 
The local coordinate frame of each quadrotor is adjusted so that its $z$-axis is along the negative gravitational force direction that is measured by the onboard IMU sensor.
Note that \eqref{eq:Aij} can be considered as a rotation matrix about the $z$-axis and a scaling along the $z$-direction of each quadrotor's local coordinate frame. Hence, \eqref{eq:HolonomCtrl} can be implemented using only the local relative position measurements.

We assume that the low-level flight controllers of the quadrotors regulate their linear velocities to the desired values $u_i$ given by \eqref{eq:HolonomCtrl}, and the orientation of quadrotors (including yaw motion) is not considered. 
We point out that \eqref{eq:HolonomCtrl} is robust to unmodeled dynamics and disturbances, and can be extended to incorporate quadrotor dynamics and directly control propellers' thrusts (see our previous work \cite{Fathian2018b}.)

If the $A_{ij}$ matrices are chosen properly, it can be shown that under control \eqref{eq:HolonomCtrl} the desired formation is achieved up to a translation in $\br^3$, a rotation and scale factor along the $z$-axis, and a non-negative overall scale factor. To fix the scale factor, \eqref{eq:HolonomCtrl} can be augmented by a bounded, odd, and smooth map $f: \mathbb{R} \rightarrow \mathbb{R}$ as
%
\begin{equation} \label{eq:HolonomCtrlAugment}
u_i := \sum_{j \in \mathcal{N}_i}{A_{ij} \, q_j^{\,i}  + f(d_{ij} - d_{ij}^*)\, q_j^{\,i}},
\end{equation}
%
where $d_{ij} := \|q_j^{\,i}\|$ denotes the distance between quadrotors $i$ and $j$, and $d_{ij}^* \in \mathbb{R}$ is its desired value. Possible choices for $f$ are $f: x \mapsto \frac{1}{k}\arctan(x)$ or $f: x \mapsto \frac{1}{k}\tanh(x)$, where $k > 0$ is an arbitrary constant. The role of $f$ in \eqref{eq:HolonomCtrlAugment} is to push quadrotors toward/away from their neighbors to achieve the desired scale.

\subsection{Control Gain Design} 

To design gain matrices that lead to desired formation, we propose a novel approach based on a semidefinite programming (SDP) formulation. Our approach is described in Algorithm \ref{alg:GainDesign}, where 
$A  \in \br^{3n \times 3n}$ is the aggregate gain matrix consisting of the $A_{ij}$ blocks, $\lambda_1(\cdot)$ denotes the smallest eigenvalue of a matrix, and
$q^*, \, \bar{q}^*,\,  \bbar{q}^* \in \br^{3n}$ respectively denote the aggregate vectors of desired formation coordinates, $90$ degree rotated coordinates about the $z$-axis, and projected coordinates on the $x$-$y$ plane. Further, $\mathbf{1}_x := \mathbf{1}_n \otimes [1,\, 0,\, 0]^\top$, $\mathbf{1}_y := \mathbf{1}_n \otimes [0,\, 1,\, 0]^\top$, $\mathbf{1}_z := \mathbf{1}_n \otimes [0,\, 0,\, 1]^\top$, where $\mathbf{1}_n \in \br^n$ is the vector of $1$'s, and $\otimes$ denotes the Kronecker product.

It can be shown that if the sensing graph among quadrotors is undirected and universally rigid, Algorithm \ref{alg:GainDesign} is guaranteed to find the gain matrices that bring the quadrotors to the desired formation. 
If the sensing graph is time-varying, by adding additional constraints to the SDP problem one can ensure that the desired formation is achieved regardless of the switches in the sensing topology. This idea has been explained in more detail in our previous work \cite{Fathian2017}.

\begin{algorithm}[t] 
%
\DontPrintSemicolon
\SetKwData{Left}{left}\SetKwData{This}{this}\SetKwData{Up}{up}
\SetKwFunction{Union}{Union}\SetKwFunction{FindCompress}{FindCompress}
\SetKwInOut{Input}{input}\SetKwInOut{Output}{output}

\SetKwInput{StepA}{step 1}
\SetKwInput{StepB}{step 2}
\SetKwInput{StepC}{step 3}
\SetKwInput{StepD}{step 4}
\SetKwInput{Notation}{notation}

\caption{Control gain design.}

\Input{Desired formation coordinates $q^*$.}
\Output{Aggregate gain matrix $A$.}

\BlankLine

\StepA{Let $N := [q^*,\, \bar{q}^*,\, \bbar{q}^*,\, \mathbf{1}_x,\, \mathbf{1}_y, \, \mathbf{1}_z]$.} 
\StepB{Compute SVD of $N = U\, S \, V^\top$.}
\StepC{Define $Q$ as the last $3n- 6$ columns of $U$.}
\StepD{Using a SDP solver, solve 
%
\begin{align} \label{eq:OptimCVX}
A ~=~ \underset{a_{ij},\, b_{ij},\, c_{ij}}{\mathrm{argmax}}~~~& \quad \lambda_1 (-Q^\top \, A \, Q)   \\
\text{subject to}& \quad A \, N = 0  \nonumber 
\end{align}
%
}
%
\label{alg:GainDesign}
\end{algorithm}


\subsection{Properties} 

By using the gains computed from Algorithm~\ref{alg:GainDesign} in \eqref{eq:HolonomCtrl} (or \eqref{eq:HolonomCtrlAugment}), it can be shown that the following properties hold:
\begin{itemize}
	\item The control is robust to perturbations, noise, and disturbances in the position measurements.
	\item The control is robust to unmodeled dynamics and input saturations.
	\item Convergence to the desired formation is unaffected by any positive scaling of the control vector.
	\item Convergence to the desired formation is unaffected by any rotation of control vector up to $90$ degrees.
\end{itemize}	

The last property indicates that the control is robust to any inaccuracies in the measured direction of gravitational force, which can be caused by noise or quadrotor acceleration. This is because such effects can be modeled as a small rotation in the desired direction of motion, for which convergence to the desired formation is not affected. 
This property can further be exploited to design a fully distributed collision avoidance scheme. Especially, each quadrotor is encapsulated in a safety cylinder to avoid both collisions and air flow disturbances. If the desired control direction points toward the safety region of another vehicle, the control is rotated to point outside of this region. If the required rotation is above $90$ degrees, the quadrotor stops until a feasible direction becomes available. Although gridlocks can occur due to the distributed nature of this strategy, in our simulations we have observed that if quadrotors are initially far apart, they can often overcome them and converge to the desired formation. 
%

\section{Experimental Results} \label{sec:Experiments}

\begin{figure*}[t!]
	\centering
	\begin{subfigure}[b]{1.0\textwidth}
		\includegraphics[trim = 1mm 54mm 3mm 53mm, clip, width=1.0\textwidth] {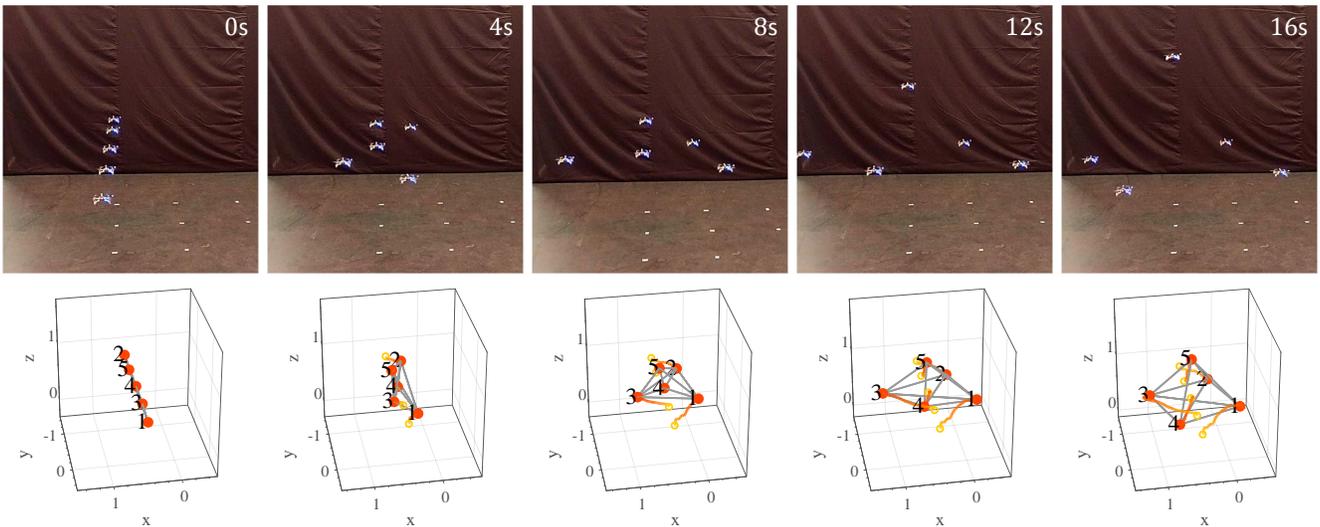}
	\end{subfigure}%
	\caption{(Top) Snapshots of experiment video for 5 quadrotors with a square pyramid desired formation. (Bottom) Coordinates of quadrotors associated to each snapshot plotted using Vicon data.}
	\label{fig:Snaps}	
\end{figure*}

We present an experiment to show that a team of quadrotors can achieve a desired formation without collision. Our experiment is based on Crazyflie quadrotors, shown in Fig.~\ref{fig:Quads}, and Vicon motion capture system for localization.
Although the position feedback provided by the Vicon system is given in a global coordinate frame, only the relative position measurements of neighboring vehicles are used to compute the control direction for each quadrotor. Furthermore, to mimic the direction of gravitational force measured by an IMU sensor, the $z$-axes of quadrotors' body frames returned by the Vicon system are chosen as the direction of gravity. Note that in general, the $z$-axes are not perfectly aligned as the quadrotors move to achieve the desired formation. This effectively demonstrates the robustness of our method to measurement inaccuracies in the direction of gravitational force that may be present in a fully distributed implementation. 

Fig.~\ref{fig:Snaps} shows snapshots of the experiment for 5 quadrotors while reaching a desired square pyramid formation.
The sensing topology among the quadrotors is chosen as a complete graph.
The quadrotors are initially placed on a line on the ground. They then move to an altitude of 70 centimeters and execute the control stratey \eqref{eq:HolonomCtrlAugment} to achieve the desired formation. 
The location and trajectory of quadrotors constructed using Vicon measurements are shown below the associated snapshots, where the sensing topology among quadrotors is shown by gray lines connecting them. 
As can be seen from the figure, the quadrotors achieve the desired formation, where by rotating their desired direction of motion when necessary they avoid collision.
Link to experiment videos are provided in the Supplementary Material section.

\section{Future Work} \label{sec:Conclusion}

Due to the distributed nature of our collision avoidance strategy gridlocks can occur. Incorporating a communication scheme such that agents can detect/avoid the gridlock situations will be a topic of future work.
Further future research include leveraging leader-follower strategies for cooperative navigation of multiple UAVs, hence, allowing a single human operator to navigate a team of UAVs while they autonomously travel in a specified formation.

\bibliographystyle{IEEEtran}
\bibliography{msBibs}

\begin{thebibliography}{10}
\providecommand{\url}[1]{#1}
\csname url@samestyle\endcsname
\providecommand{\newblock}{\relax}
\providecommand{\bibinfo}[2]{#2}
\providecommand{\BIBentrySTDinterwordspacing}{\spaceskip=0pt\relax}
\providecommand{\BIBentryALTinterwordstretchfactor}{4}
\providecommand{\BIBentryALTinterwordspacing}{\spaceskip=\fontdimen2\font plus
\BIBentryALTinterwordstretchfactor\fontdimen3\font minus
  \fontdimen4\font\relax}
\providecommand{\BIBforeignlanguage}[2]{{%
\expandafter\ifx\csname l@#1\endcsname\relax
\typeout{** WARNING: IEEEtran.bst: No hyphenation pattern has been}%
\typeout{** loaded for the language `#1'. Using the pattern for}%
\typeout{** the default language instead.}%
\else
\language=\csname l@#1\endcsname
\fi
#2}}
\providecommand{\BIBdecl}{\relax}
\BIBdecl

\bibitem{Cieslewski2018}
T.~Cieslewski, S.~Choudhary, and D.~Scaramuzza, ``Data-efficient decentralized
  visual {SLAM},'' in \emph{IEEE International Conference on Robotics and
  Automation}, 2018.

\bibitem{Ozaslan2017}
T.~Ozaslan, G.~Loianno, J.~Keller, C.~Taylor, V.~Kumar, J.~Wozencraft, and
  T.~Hood, ``Autonomous navigation and mapping for inspection of penstocks and
  tunnels with {MAV}s,'' \emph{IEEE Robotics and Automation Letters}, vol.~2,
  no.~3, pp. 1740--1747, 2017.

\bibitem{Dorling2017}
K.~Dorling, J.~Heinrichs, G.~G. Messier, and S.~Magierowski, ``Vehicle routing
  problems for drone delivery,'' \emph{IEEE Transactions on Systems, Man, and
  Cybernetics: Systems}, vol.~47, no.~1, pp. 70--85, 2017.

\bibitem{Baehnemann2017}
R.~Bähnemann, D.~Schindler, M.~Kamel, R.~Siegwart, and J.~Nieto, ``A
  decentralized multi-agent unmanned aerial system to search, pick up, and
  relocate objects,'' in \emph{IEEE International Symposium on Safety, Security
  and Rescue Robotics}, Oct 2017, pp. 123--128.

\bibitem{Oh2015}
K.-K. Oh, M.-C. Park, and H.-S. Ahn, ``A survey of multi-agent formation
  control,'' \emph{Automatica}, vol.~53, pp. 424--440, 2015.

\bibitem{Michael2008}
N.~Michael, M.~M. Zavlanos, V.~Kumar, and G.~J. Pappas, ``Distributed
  multi-robot task assignment and formation control,'' in \emph{IEEE
  International Conference on Robotics and Automation}, 2008, pp. 128--133.

\bibitem{Aranda2015}
M.~Aranda, G.~L{\'o}pez-Nicol{\'a}s, C.~Sag{\"u}{\'e}s, and M.~M. Zavlanos,
  ``Coordinate-free formation stabilization based on relative position
  measurements,'' \emph{Automatica}, vol.~57, pp. 11--20, 2015.

\bibitem{Hyun2016}
N.-s.~P. Hyun, P.~A. Vela, and E.~I. Verriest, ``Collision free and permutation
  invariant formation control using the root locus principle,'' in
  \emph{American Control Conference}, 2016, pp. 2572--2577.

\bibitem{Trinh2018}
M.~H. Trinh, S.~Zhao, Z.~Sun, D.~Zelazo, B.~D. Anderson, and H.-S. Ahn,
  ``Bearing-based formation control of a group of agents with leader-first
  follower structure,'' \emph{IEEE Transactions on Automatic Control}, 2018.

\bibitem{Fathian2016}
K.~Fathian, D.~I. Rachinskii, M.~W. Spong, and N.~R. Gans, ``Globally
  asymptotically stable distributed control for distance and bearing based
  multi-agent formations,'' in \emph{American Control Conference}, 2016, pp.
  4642--4648.

\bibitem{Fathian2016a}
K.~Fathian, D.~I. Rachinskii, T.~H. Summers, and N.~R. Gans, ``Distributed
  control of cyclic formations with local relative position measurements,'' in
  \emph{IEEE Conference on Decision and Control}, 2016, pp. 49--56.

\bibitem{Han2018}
T.~Han, Z.~Lin, R.~Zheng, and M.~Fu, ``A barycentric coordinate-based approach
  to formation control under directed and switching sensing graphs,''
  \emph{IEEE Transactions on cybernetics}, vol.~48, no.~4, pp. 1202--1215,
  2018.

\bibitem{Fathian2018}
K.~Fathian, T.~H. Summers, and N.~R. Gans, ``Robust distributed formation
  control of agents with higher-order dynamics,'' \emph{Control Systems
  Letters}, 2018.

\bibitem{Fathian2018b}
K.~Fathian, S.~Safaoui, T.~H. Summers, and N.~R. Gans, ``Robust distributed
  planar formation control for higher-order holonomic and nonholonomic
  agents,'' \emph{arXiv preprint, arXiv:1807.11058}, 2018.

\bibitem{Leishman2014}
R.~C. Leishman, J.~Macdonald, R.~W. Beard, and T.~W. McLain, ``Quadrotors and
  accelerometers: State estimation with an improved dynamic model,'' \emph{IEEE
  Control Systems Magazine}, 2014.

\bibitem{Fathian2017}
K.~Fathian, D.~I. Rachinskii, T.~H. Summers, M.~W. Spong, and N.~R. Gans,
  ``Distributed formation control under arbitrarily changing topology,'' in
  \emph{American Control Conference}, 2017, pp. 271--278.

\end{thebibliography}

\end{document}